\documentclass{article}
\usepackage{ijcai24}

\usepackage{times}
\usepackage{soul}
\usepackage{url}
\usepackage[hidelinks]{hyperref}
\usepackage[utf8]{inputenc}
\usepackage[small]{caption}
\usepackage{graphicx}
\usepackage{amsmath}
\usepackage{amsthm}
\usepackage{booktabs}
\usepackage{algorithm}
\usepackage{algorithmic}
\usepackage[switch]{lineno}

\usepackage{inconsolata}
\usepackage{graphicx}
\usepackage{adjustbox}
\usepackage{color}
\usepackage{tabularx}
\usepackage{comment}
\usepackage{amssymb}
\usepackage{bm}
\usepackage{mathtools}

\DeclarePairedDelimiter\floor{\lfloor}{\rfloor}
\usepackage{multirow}
\usepackage{makecell}
\usepackage{caption}
\usepackage{subcaption}
\usepackage{tablefootnote}

\newcommand{\bing}[1]{\textcolor{blue}{[Bing: #1]}}

\title{Many Hands Make Light Work: Task-Oriented Dialogue System with Module-Based Mixture-of-Experts}

\author{
\begin{tabular}[t]{cc} 
    \author{
    Ruolin Su  \and Biing-Hwang Juang\\
    \affiliations
    Georgia Institute of Technology\\
    \emails
    ruolinsu@gatech.edu \and juang@ece.gatech.edu}\\
\end{tabular}
}

\begin{document}

\maketitle

\begin{abstract}
    Task-oriented dialogue systems are broadly used in virtual assistants and other automated services, providing interfaces between users and machines to facilitate specific tasks.
    Nowadays, task-oriented dialogue systems have greatly benefited from pre-trained language models (PLMs).
    However, their task-solving performance is constrained by the inherent capacities of PLMs, and scaling these models is expensive and complex as the model size becomes larger. To address these challenges, we propose \textbf{S}oft \textbf{M}ixture-of-\textbf{E}xpert \textbf{T}ask-\textbf{O}riented \textbf{D}ialogue system (\textbf{SMETOD}) which leverages an ensemble of Mixture-of-Experts (MoEs) to excel at subproblems and generate specialized outputs for task-oriented dialogues. SMETOD also scales up a task-oriented dialogue system with simplicity and flexibility while maintaining inference efficiency. We extensively evaluate our model on three benchmark functionalities: intent prediction, dialogue state tracking, and dialogue response generation. Experimental results demonstrate that SMETOD achieves state-of-the-art performance on most evaluated metrics. Moreover, comparisons against existing strong baselines show that SMETOD has a great advantage in the cost of inference and correctness in problem-solving.
\end{abstract}

\section{Introduction}

Task-oriented dialogue systems play a crucial role in virtual assistants and various automated services through human-machine interactions.
The fundamental objective of a task-oriented dialogue system is to aid users in completing specific services or tasks all achieved through natural language dialogues~\cite{wen-etal-2017-network}.
Considering a broad range of applications, task-oriented dialogue systems should generate diverse types of outputs for processing information, evaluating user intentions, or retaining for future reference.
In real-world scenarios, useful information processed from dialogue could be presented in various formats, including form-based~\cite{Goddeau1996form}, probability-based~\cite{thomson2010bayesian}, or text-based~\cite{hosseini2020simple}.
Typically, several components are responsible for managing a variety of information: natural language understanding (\textbf{NLU}) for comprehending and translating user intent into either natural language or a format suitable for machine processing, dialogue state tracking (\textbf{DST}) for discerning the user's requirements and providing a foundation for subsequent decisions, and natural language generation (\textbf{NLG}) generate a natural language response to the user based on the machine’s decision of the next move.

This leads to two predominant system designs, namely pipeline-based and end-to-end, divided by whether the machine-generated response is based on dialogue utterances or processed information from other components only.
Either system design presents its own set of limitations in effectively addressing diverse output objectives~\cite{takanobu2020your}.
Drawbacks of pipeline-based systems lie in the potential for error propagation from one module to another, and local decisions can have adverse global effects~\cite{su2016continuously}.
End-to-end dialogue systems, on the other hand, raise concerns about missing all essential information that may be required other than responses. Moreover, diagnosing and considering component-flow characteristics can be challenging in end-to-end systems~\cite{bang2023task}.

 \begin{figure*}[htb]
  \centering
  \includegraphics[width=.95\textwidth]{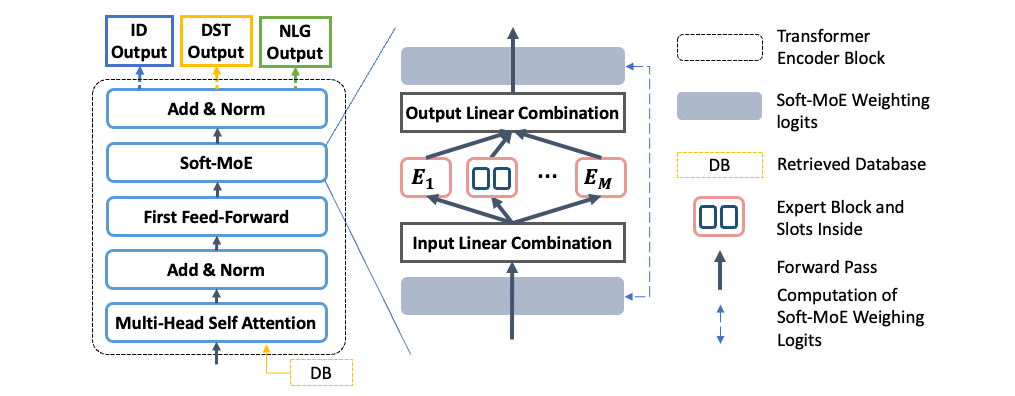}
  \caption{Architecture of the \textbf{SMETOD} as in Transformer encoders. The result from the DB derived from the output of DST is used for NLG inference. All of the expert layers share the same architecture. The input is ensembled by experts in the Soft-MoE layer for improving model capacity without the cost of efficiency. The model is fine-tuned by maximizing the likelihood of predicting the next token for NLU, DST, and NLG outputs.}
  \label{fig:model}
\end{figure*}

Despite the limitations in dialogue-system designs, there are also significant constraints in terms of scaling dialogue models with efficiency.
Recent advancements have leveraged the transfer learning capabilities of pre-trained language models (PLMs)~\cite{radford2019language,raffel2020exploring} by fine-tuning~\cite{budzianowski2019hello,heck2020trippy} or pre-training dialogue models~\cite{wu-etal-2020-tod,zhang-etal-2020-dialogpt}. 
However, their remarkable performance is at the cost of significant computational resources, especially as the sizes of PLMs continue to grow.
Recently, parameter-efficient adapters raised that freeze the PLM while only allowing a small number of parameters updated for downstream models~\cite{houlsby2019parameter}, and have gained popularity in dialogue systems~\cite{bang2023task}. 
Nevertheless, the model capacity (i.e. number of parameters) is limited by the number of downstream models, and the addition of adapters can become computationally expensive due to their sequential processing~\cite{ruckle2020adapterdrop}.
We also argue that the issue of inference time scaling with model complexity becomes more prominent considering the time sensitivity associated with the deployment of dialogue systems.

To address these issues, we propose a \textbf{S}oft \textbf{M}ixture-of-\textbf{E}xpert \textbf{T}ask-\textbf{O}riented \textbf{D}ialogue system (\textbf{SMETOD}) which scales the model capacity for diverse outputs of dialogue systems with significantly less training and inference cost. 
Specifically, we leverage Soft MoE~\cite{puigcerver2023sparse} to improve model capacity and leverage the effectiveness and performance of considerably larger models with significantly lower computational costs.
We present a task-oriented dialogue system as a multi-module end-to-end text generation to bridge the gap between traditional pipeline-based and end-to-end response generation systems, and optimize NLU, DST, and NLG, respectively, as in~\cite{su2021multitask,bang2023task}.
We formulate NLU, DST, and NLG as the text generation problems, which take dialogue history sequence as model input and generate spans as the output. 
In the cases of NLG, we predict the DST output to obtain the database (DB) state, which becomes incorporated into its input.
With T5-small and T5-base~\cite{2020t5} as the backbone PLM, we evaluate our method on MultiWOZ~\cite{eric2019multiwoz,zang2020multiwoz} and NLU ~\cite{casanueva2020efficient,larson-etal-2019-evaluation,liu2019benchmarking} datasets. 

Our contribution is as follows:
\begin{itemize}
\item We propose SMETOD, a task-oriented dialogue system for diverse outputs, which first leverages Soft-MoE in dialogue systems to improve model capacity with efficiency for specialized output generation.
\item Experimental results demonstrate the effectiveness of our model by achieving SOTA generative performance of NLU and DST on all evaluation benchmarks and improving most evaluation metrics on NLG.
\item Our study of time efficiency and the architect of Soft-MoE proves the significant improvement of efficiency as model complexity continues to grow, promoting future studies on dialogue system design with efficiency.

\end{itemize}

\section{Preliminaries}

\noindent \textbf{Soft Mixture-of-Experts.} 
Mixture-of-Experts (MoE)-based models have shown advantages in scaling model capacity without large increases in training or inference costs. 
There has been work on scaling sparsely activated MoE architectures. 
In the context of modern deep learning architectures, it was firstly found effective by stacking MoE between the LSTM and resulted in the state-of-the-art in language modeling and machine translation~\cite{shazeer2017}. Later on, the MoE Transformer was introduced where MoE layers are a substitute for the FFN layers \cite{shazeer2018mesh}. 

We adopt Soft-MoE~\cite{puigcerver2023sparse}, which scales model capacity without the loss of fine-tuning efficiency and is fully differentiable and balanced compared to conventional efficient MoEs, for example, GShard~\cite{lepikhin2020gshard}. Specifically, it performs a soft assignment on experts to each input token, achieving similar training costs and much lower inference costs at a larger model capacity.
We use $f(\cdot;\theta)$ to denote a mapping $f$ associated with the parameter $\theta$ from the input sample to an output space. $\sigma(\cdot)$ is the Softmax function. 
We denote $\{f(\cdot;\theta_i)\}_{i=1}^m$ as \textbf{$\bm{m}$ experts} with identical architectures; their weights $\theta_1, \dots, \theta_m$ are applied to individual tokens. 
Each expert has \textbf{$\bm{p}$ slots}, each of which is a weighted average of input values. Slots in the same expert have the same weights.
Given input and output tokens $\bm{x}=\{x_1, \dots, x_l\}$ and $\bm{y}=\{y_1, \dots, y_l\}$ at the length $l$. Each expert will process $p$ slots with parameters denoted as $\Psi = \{\psi^{(1)}, \dots, \psi^{(m \times p)}\}$.
The input of experts, $\tilde{\bm{x}}$, is defined as the result of convex combinations of input tokens.
\begin{equation}
    \label{dispatch}
    \tilde{\bm{x}_j} = (\sigma(\bm{x}\psi^{(j)}))^T \bm{x}
\end{equation}
where 
$j$ is the index of the slot in experts and $j\in[1,\dots,m\times p]$. The corresponding expert function is applied on each slot to obtain the output slots:
\begin{equation}
    \label{expert-layer}
    \tilde{\bm{y}_j} = f( \tilde{\bm{x}_j} ; \theta_{\floor*{j/p}})
\end{equation}
Given $\tilde{\bm{y}}=\{\tilde{\bm{y}_j}\}_{j=1}^{m\times p}$, the output of Soft-MoE layer, $y_i$, is computed as a convex combination of all $(m \times p)$ output slots over the expert dimension (i.e. the rows of $\bm{x}\Psi$):
\begin{equation}
    \label{combine}
    y_i = \sigma(x_i\Psi)~\tilde{\bm{y}}
\end{equation}

\noindent \textbf{End-to-end task-oriented dialogue system.}
End-to-end learning was found effective in training and optimizing the map directly from input to output~\cite{wen-etal-2017-network,liu2018end,eric2017copy,williams2017hybrid}. 
Later on, a lot of endeavor was given to fine-tuning pre-trained language models and adapting their generalization capacities for an end-to-end system of task-oriented dialogues~\cite{budzianowski2019hello,hosseini2020simple}. 
In recent years, pre-trained task-oriented dialogue models have emerged as strong contenders, surpassing traditional fine-tuning approaches and showcasing competitive generalization capabilities, particularly in multi-objective scenarios~\cite{wu-etal-2020-tod,zhang-etal-2020-dialogpt,peng2021soloist,he2022unified}. 
However, it's worth noting that they require a large amount of dialogue data to train the backbone models and without an interface to optimize sub-modules.

\noindent \textbf{Efficient transfer learning.}
To reduce the effort in tuning large PLMs and promote the scalability of model adaptation,
there is a line of work that fixes the entire PLM and introduces a small number of new trainable parameters. Notable examples in this category include adapters~\cite{houlsby2019parameter}, prefix-tuning~\cite{li2021prefix} and prompt-tuning~\cite{lester2021power}, \textit{etc}.
In-context learning prepends related task examples to condition on the generated dialogue states~\cite{hu2022context}. 
In end-to-end dialogue systems, a line of work prompts with specific text to generate desired outputs~\cite{su2021multitask} or injecting adapters to capture the knowledge of different functionalities~\cite{bang2023task,mo2023parameter}.
GPT-3~\cite{brown2020language} and ChatGPT\footnote{\url{https://chat.openai.com/chat}} are also successful and efficient open-domain dialogue systems.
On the other hand, the MoE approach focuses on improving performance by efficiently scaling model sizes. 

\section{Method}
 We introduce \textbf{SMETOD}, a multi-objective dialogue system for NLU, DST, and NLG in task-oriented dialogues, scaling model capacities while maintaining computational efficiency with Soft MoE. The overall architecture is illustrated in Figure~\ref{fig:model}.

 \subsection{Problem Formulation}
 We define the dialogue history $h = \left[u_1^{sys},u_1^{usr},\dots,u_t^{sys},u_t^{usr}\right]$ as the concatenation of the system and user utterances in previous turns, where $t$ is the number of current turns in the dialogue.
 $h$ has all the dialogue history without the last system utterance, denoted as $r$. 
 NLU outputs an $I$ which is an intent or the API-name.
 The objective of DST is to output user goals, the tasks or purposes that the user wants to accomplish through the dialogue. user goals are typically represented as a set of pre-defined slot-value pairs that consist of the required information to query the dialogue system, i.e. $y_{API} = \{(s_1,v_1), \dots, (s_{n},v_{n})\}$, where $n$ is the number of slot-value pairs.
 Finally, NLG will generate $S$ with the previous output: $h+y_{DB} \rightarrow r$, where $y_{DB}$ is the items in the database retrieved by $y_{API}$.
 Given a pair of training examples $(x', y')$, we elaborate $x'$ and $y'$ corresponding to different modules of the dialogue system in the following Table.
 
 \begin{table}[h]
    \centering
    \begin{tabular}{ccc}
    \toprule
         &  $x'$ & $y'$\\
\hline
 NLU& $h$ & $I$\\
 DST& $h$ & $y_{API}$\\
 NLG& $h + y_{DB}$& $r$\\

    \bottomrule
    \end{tabular}
    \label{tab:annotation}
\end{table}

 \subsection{Soft Mixture-of-Expert Layer}
 We implement the Soft-MoE layer to replace the second Feed-Forward Layer in each Transformer~\cite{vaswani2017attention} Encoder block, as illustrated in Figure~\ref{fig:model}. 
 Mathematically, we denote the output out the first Feed-Forward layer of the $k$-the encoder is $g(\cdot;\phi_k)$, then $ \bm{x} = g(x';\phi_k) \in \mathbb{R}^{l \times d_{ff}}$ in Eq.~\ref{dispatch}, denoting $d_{ff}$ as the dimension between the first and second Feed-Forward layer and $d$ as model's hidden dimension, and $l$ is the length of tokens.
  $\psi^{(j)} \in \mathbb{R}^{d_{ff}}$ is $d_{ff}$-dimensional vector of parameters corresponding to each slot of experts.
 
 The mapping $f(\cdot;\theta_i)$ in Eq.~\ref{expert-layer} is simply a linear mapping corresponding to each expert, and $p$ is the slots per expert having the same weights.
 Therefore, the output of the $k$-th encoder layer, $y'^{(k)}$, can be represented as 
 \begin{equation}
\begin{split}
     y'^{(k)} = f(g(x';\phi_k); \Theta_k, \Psi_k)
 \end{split}
 \end{equation}
 
 For fine-tuning, we replicate the pre-trained weights from the second Feed-Forward layer of encoders and assign them to each expert, leveraging the contextual learning abilities inherent in pre-trained models.

 \subsection{Training Objectives}

 We optimize the generation outputs of NLU, DST, NLG, respectively, following~\cite{su2021multitask}. 
 Given a pair of training samples as $(x', y')$,  the loss function is defined to maximize the log-likelihood of the token to predict given the current context:
  \begin{equation}
     \mathcal{L}_{\{NLU, DST, NLG\}} = -\frac1l \sum_{q=1}^l \log P(y'_q|y'_{<q};x')
 \end{equation}

\section{Experiment}
\subsection{Data}
We evaluate our models for NLU on Banking77~\cite{casanueva2020efficient}, CLINC150~\cite{larson-etal-2019-evaluation}, and HWU64~\cite{liu2019benchmarking}; DST and NLG are evaluated on the task-oriented dialogue benchmarks MultiWOZ 2.1~\cite{eric2019multiwoz} and MultiWOZ 2.2~\cite{zang2020multiwoz}.
Banking77 contains 13,083 customer service queries labeled with 77 distinct intents for distinguishing between intents among queries related to similar tasks.
CLINC150, consists of a comprehensive dataset comprising 23,700 examples, annotated with 150 intents across 10 distinct domains.
HWU64 is collected from the home robot that has 25,716 examples for 64 intents spanning 21 domains.

MultiWOZ 2.1~\cite{eric2019multiwoz} consists of multi-turn task-oriented dialogues across several domains, where 8,438 dialogues are for training and 1,0000 for dev and test.
MultiWOZ 2.2~\cite{zang2020multiwoz} improves MultiWOZ 2.1 by correcting annotation errors and adding dialogue act annotations.
In MultiWOZ, the generation of response is not only related to the dialogue context but also grounded on the database (DB) state.
The DB state is automatically retrieved from a pre-defined database using the generated dialogue states.
SMETOD adopts a two-step approach during inference. Firstly, it predicts the DST results to access the DB state. Subsequently, it utilizes the retrieved DB state and the current dialogue context to generate the NLG results.

\begin{table}[t]
    \centering
    \begin{adjustbox}{width=\linewidth}
    \begin{tabular}{cccc}
    \toprule
         Model &Banking77&  HWU64& CLINC150\\
         \midrule
         \makecell{BERT{\small-FIXED}$^\diamond$$^*$ }& 87.19& 85.77& 91.79\\
        \makecell{CONVBERT-DG\\{\small+Pre+Multi}$^*$}& 92.99& 92.94&97.11\\
        \makecell{CONVBERT\\{\small+Pre+Multi}$^*$}&93.44& 92.38&97.11\\
         \makecell{BERT{\small-TUNED}$^\diamond$$^*$ }& 93.66& 92.10& 96.93\\
        \makecell{CONVERT$^*$} & 93.01 & 91.24 & 97.16 \\
        \makecell{USE+CONVERT$^*$ }& 93.36 & 92.62 & 97.16 \\
        \makecell{SPACE-2$^\sharp$$^*$}&94.77&  \textbf{94.33}$^*$& 97.80\\
         \makecell{SPACE-3$^*$ }&\textbf{94.94}$^*$&  94.14& 97.89\\
         \makecell{TOATOD$_{small}$}&92.40&  90.42& 98.45\\
         \makecell{TOATOD$_{base}$ }&92.17&  90.79& 98.01\\
         \Xhline{\arrayrulewidth}
         \makecell{SMETOD$_{small}$ }&92.47&  90.88&  98.12\\
         \makecell{SMETOD$_{base}$ }&\textbf{93.02}&  \textbf{92.56}& \textbf{98.64}\\
    \bottomrule
    \end{tabular}
    \end{adjustbox}
    \caption{Accuracy (\%) on three intent prediction datasets with full-data experiments. \# are obtained from DialoGLUE leaderboard$^2$. All others are reported as in the original papers. Models with * are classification-based. }
    \label{tab:nlu}
\end{table}

\footnotetext[2]{\url{https://eval.ai/web/challenges/challenge-page/708/leaderboard/1943}}
\subsection{Training \& Inference Details}
All models are fine-tuned respectively using PPTOD~\cite{su2021multitask}, the pre-trained dialogue models based on T5-small (60M parameters) and T5-base (220M parameters)~\cite{raffel2020exploring}, as the backbone models. 
T5-small has 6 encoders and decoders with hidden size $d=512$ and $d_{ff}=2048$. While T5-large has 12 encoders and decoders and $d=768$, $d_{ff}=3072$.
For models' architecture, we replace the second Feed-Forward layer in all encoder blocks with the illustrated Soft-MoE layers, and copy pre-trained weights to each expert in the Soft-MoE layers. We continue to train backbone models on the heterogeneous dialogue corpora similar to \cite{su2021multitask}.
We augment T5 with 8 experts and 2 slots per expert for DST, and 16 experts with 2 slots per expert for NLU and NLG.

We fine-tuned all model parameters on the full-shot training setting.
The linear combination weights in Soft-MoE layer are initialized by Kaiming initialization~\cite{He_2015_ICCV}.
The initial learning rate is set to 0.001 for NLU, and 0.0001 DST, NLG, respectively. 
We use the Adafactor~\cite{shazeer2018adafactor} optimizer and the training batch size is set to 64 on Nvidia A10 GPUs.
We tried a wide range of learning rates from 1e-2 to 1e-6 then set the initial training rate to 1e-4 in all training.
Our code is developed based on \textit{Soft-Mixture-of-Experts}\footnotemark[3]\footnotetext[3]{\url{https://github.com/fkodom/soft-mixture-of-experts.git}} and \textit{TOATOD}\footnotemark[4]\footnotetext[4]{\url{https://github.com/sogang-isds/TOATOD.git}}. Code repository will be released to the public soon.

Because different batch sizes will result in different padded lengths, inference results are slightly changed by batch sizes due to Softmax over input tokens in the Soft-MoE layer. We make inferences on several selected batch sizes and report average scores. We found out that different batch sizes in our experiments have negligible influence on the inference results\footnotemark[5]\footnotetext[5]{We conducted a hypothesis test and found out p-value $<0.01$ for scores changed by batch size. }. 

\begin{table*}[h]
    \centering
    \begin{adjustbox}{width=.6\linewidth}
    \begin{tabular}{cccc}
    \toprule
         Model& Pre-Trained Model & MultiWOZ2.1&MultiWOZ2.2\\
\toprule
 TRADE&- &45.6&45.4\\
 \Xhline{.2\arrayrulewidth}
 TripPy& BERT-base&55.29&-\\
 TripPy$_{+SaCLog}$& BERT-base&60.61&-\\
 CONVBERT-DG& BERT-base&55.29&-\\
 \Xhline{.2\arrayrulewidth}
 SimpleTOD& DistilGPT-2 & 55.76&-\\
 SOLOIST& GPT-2 & 56.85&-\\
 \Xhline{.2\arrayrulewidth}
 AG-DST& PLATO-2 & 57.26& 57.26\\
 \Xhline{.2\arrayrulewidth}
 UniLM$^\ddag$& UniLM& 54.25& 54.25\\
 SPACE-3& UniLM& 57.50& 57.50 \\
 \Xhline{.2\arrayrulewidth}
 PPTOD$_{base}$& T5-base& 57.10&-\\
 PPTOD$_{large}$& T5-large& 57.45&-\\
 D3ST$_{base}$& T5-base &54.2 & 56.1\\
 D3ST$_{large}$& T5-large & 54.5&54.2\\
 D3ST$_{XXL}$& T5-XXL & 57.80&58.7\\
 T5DST$_{+desc}$ & T5-base& 56.66 & 57.6\\
 TOATOD$_{small}$ $^\dag$& T5-small& 59.49& 59.33\\
 TOATOD$_{base}$ $^\dag$& T5-base&  59.51& 60.02\\
 \midrule
 \midrule
 SMETOD$_{small}$& T5-small&  59.69& 59.60\\
 SMETOD$_{base}$& T5-base&  \textbf{60.36}& \textbf{60.08}\\
    \bottomrule
    \end{tabular}
    \end{adjustbox}
    \caption{Joint Goal Accuracy (\%)  for DST on MultiWOZ 2.1 and 2.2. Results with \ddag~are from SPACE-3 paper. \dag~represents the results of our re-implementation. All others are reported as in the original papers.}
    \label{tab:dst}
\end{table*}
 
\begin{table*}
    \centering
    \begin{adjustbox}{width=.9\linewidth}
    \begin{tabular}{cccccccccc}
    \toprule
    \multirow{2}{*}{\textbf{Model}} & \multirow{2}{*}{\textbf{Backbone}} &
    \multicolumn{4}{c}{MultiWOZ2.1}&  \multicolumn{4}{c}{MultiWOZ2.2}\\
    \cmidrule(r{1\cmidrulekern}){3-6}  \cmidrule(l{1\cmidrulekern}){7-10}
    & & \textit{Inform} & \textit{Success} & \textit{BLEU} & \textit{Combined}  & \textit{Inform} & \textit{Success} & \textit{BLEU} & \textit{Combined}\\
    \midrule
    DOTS& BERT-base& 86.65& 74.18& 15.90& 96.32& -& -& -& -\\
    DiactTOD & S-BERT& -& -& -& -& 	89.5&	\textbf{84.2}&	17.5& 104.4\\
    \Xhline{.2\arrayrulewidth}
    SimpleTOD& DistilGPT-2 &85.00& 70.50& 15.23 &92.98& -& -& -& -\\
    SOLOIST& GPT-2& -& -& -& -& 82.3& 72.4& 13.6& 90.9\\
     UBAR$^\vartriangle$& GPT-2 & 95.70 &81.80 &16.50 &105.25 &83.4 &70.3 &17.6 &94.4\\
    \Xhline{.2\arrayrulewidth}
    MinTL$^\vartriangle$ & BART$_{large}$& -& -& -& -& 73.7& 65.4& 19.4& 89.0\\
    RewardNet$^\vartriangle$& BART$_{large}$& -& -& -& -& 87.6&	81.5& 17.6&	102.2\\
    \Xhline{.2\arrayrulewidth}
    GALAXY& UniLM & 95.30& 86.20& \textbf{20.01}& \textbf{110.76} & 85.4& 75.7& \textbf{19.64}& 100.2\\
    \Xhline{.2\arrayrulewidth}
    PPTOD$_{base}$& T5-base& 87.09& 79.08& 19.17& 102.26 & -& -& -& -\\
    MTTOD$^\natural$& T5-base& 90.99& 82.08& 19.68& 106.22& 85.9& 76.5& 19.0& 100.2\\
    RSTOD$^\natural$& T5-small& 93.50&   84.70& 19.24&   108.34& 83.5& 75.0& 18.0& 97.3\\   
    TOATOD$_{small}$&  T5-small&  92.10&  80.40&  18.29&  104.54&  85.80& 74.00& 18.00& 97.90\\
    TOATOD$_{base}$&  T5-base&  97.00&  87.40&  17.12& 109.32&  90.00& 79.80& 17.04& 101.94 \\
    KRLS& T5-base & -& -& -& -& 89.2& 80.3& 19.0&	103.8\\
    \midrule
    \midrule
     SMETOD$_{small}$& T5-small& 91.80& 79.30& 16.58& 102.13& 89.7& 75.3& 16.9& 99.4\\
     SMETOD$_{base}$& T5-base& \textbf{98.08}& \textbf{87.70}& 17.18& 110.07& \textbf{91.2}& 81.7& 18.2& \textbf{104.7}\\
     \bottomrule
 
    \end{tabular}
    \end{adjustbox}
    \caption{Evaluation of NLG on Inform, Success, BLEU, and Combined Scores, where Combined = (Inform + Success) × 0.5 + BLEU. $\natural$ represents the NLG results on MultiWOZ 2.1 is borrowed from RSTOD paper. All other results are from MultiWOZ leaderboards$^6$. $\vartriangle$ shows models that require oracle dialogue states for prediction.}
    \label{tab:nlg}
\end{table*}

\section{Results \& Discussion}

We show the effectiveness of our models on \textbf{NLU} (Sec.~\ref{sec:nlu}), \textbf{DST} (Sec.~\ref{sec:dst}), and \textbf{NLG} (Sec.~\ref{sec:nlg}) in task-oriented dialogue systems compared to plenty of strong baselines. 
In the experiments, we fine-tune SMETOD using the small and base versions of PPTOD, which continues pre-training T5 on large dialogue corpora, as the start point.  
We observe that SMETOD is state-of-the-art on NLU and DST and comparable with existing baselines on NLG. 
We also study the improvement of efficiency with SMETOD (Sec.~\ref{sec:time}).
In Sec.~\ref{sec:architect}, we investigate model performance when the Soft-MoE layers are in different architectures.

\subsection{Intent Prediction}
\label{sec:nlu}

The goal of intent prediction, known as NLU in a task-oriented dialogue system, is to identify the user’s intention based on the user’s utterance. We conduct experiments on three benchmarks: Banking77~\cite{casanueva2020efficient}, CLINC150~\cite{larson-etal-2019-evaluation}, and HWU64~\cite{liu2019benchmarking}. 
We report the Accuracy (\%) of predicting an intention correctly for evaluation.

\subsubsection{Baselines}
Baselines have a wide range from BERT-based models: CONVBERT~\cite{mehri2020dialoglue}, CONVERT~\cite{casanueva2020efficient}, UniLM-based models: SPACE-2~\cite{he-etal-2022-space}, SPACE-3~\cite{he2022unified}, to T5-based TOATOD~\cite{bang2023task}.
All baseline models utilizing BERT and UniLM follow a classification-based approach, employing a classifier featuring a Softmax layer to make predictions from a predefined set of intents.

\subsubsection{Evaluation Results}
Table~\ref{tab:nlu} shows that our approaches perform state-of-art on CLINC150, which has the most number of intent types. 
On the other two benchmarks, our approaches have the highest accuracy compared to other generation-based approaches. Classification-based approaches are better which may benefit from smaller numbers of intents to choose from.
Compared to classification models, SMETOD copes with the classification task as a generation problem by directly generating the text label. 
Therefore, when adapting to a new classification task, SMETOD is more scalable to new domains and tasks and can predict intents that are not in the ontology.

\subsection{Dialogue State Tracking}
\label{sec:dst}
As a crucial component in task-oriented dialogue systems, DST determines the user goals based on the history of dialogue turns. 
For the evaluation of DST models,  we use joint goal accuracy (JGA) which is the average accuracy of predicting all slot-values for the current turn correctly.

\subsubsection{Baselines}
 In Table~\ref{tab:dst}, we compare SMETOD with a wide range of classification-based approaches: TRADE~\cite{wu2019transferable}, TripPy~\cite{heck2020trippy}, TripPy + SaCLog~\cite{dai2021preview}, CONVBERT-DG~\cite{mehri2020dialoglue}, SimpleTOD~\cite{hosseini2020simple}, SOLOIST~\cite{peng2021soloist}, AG-DST~\cite{tian2021amendable}, SPACE-3~\cite{he2022unified}, and generation-based approaches: PPTOD~\cite{su2021multitask}, D3ST~\cite{zhao2022description}, T5DST~\cite{lee-etal-2021-dialogue}, and TOATOD~\cite{bang2023task}. 

 \subsubsection{Evaluation Results}
Compared to other approaches, SMETOD obtains state-of-the-art JGA on MultiWOZ 2.1 and 2.2 among all generation-based approaches.
Our model is more flexible to generate slot-value pairs while classification-based models are limited to the pre-defined ontology.
The results show that our model can benefit from not only the transfer learning capacities of per-trained models but also the improvement of model size.

\subsection{End-to-End Response Generation}
\label{sec:nlg}
End-to-end dialogue response generation, aiming at evaluating the model in the most realistic, fully end-to-end setting, where the generated dialogue states are used for the database search and response generation~\cite{hosseini2020simple,su2021multitask}, is NLG in task-oriented dialogue system.
Our models evaluated on MultiWOZ generate responses not only related to the dialogue history but also grounded on the database (DB) state. 

\subsubsection{Metrics}
For evaluation, we follow the individual and combined metrics in \cite{hosseini2020simple}: Inform, Success, and BLEU, and Combined score which is defined as Combined = (Inform + Success) × 0.5 + BLEU. Specifically, Inform rate measures the correctness of entities in the response. Success rate success rate assesses attribute fulfillment requested by the user. BLUE score is used to measure the fluency of the generated responses.

\subsubsection{Baselines}
In Table~\ref{tab:nlg}, we compare our model with several strong baselines: DOTS~\cite{jeon2021domain}, DiactTOD~\cite{wu2023diacttod}, SimpleTOD~\cite{hosseini2020simple}, SOLOIST~\cite{peng2021soloist}, UBAR~\cite{yang2021ubar}, MinTL~\cite{lin2020mintl}, RewardNet~\cite{feng2023fantastic}, GALAXY~\cite{he2022galaxy}, PPTOD~\cite{su2021multitask}, RSTOD~\cite{cholakov2022efficient}, MTTOD~\cite{lee2021improving}, TOATOD~\cite{bang2023task}, KRLS~\cite{yu2022krls}.

\begin{figure*}[tb]
  \begin{subfigure}{.25\linewidth}
  \centering
  \includegraphics[width=\linewidth]{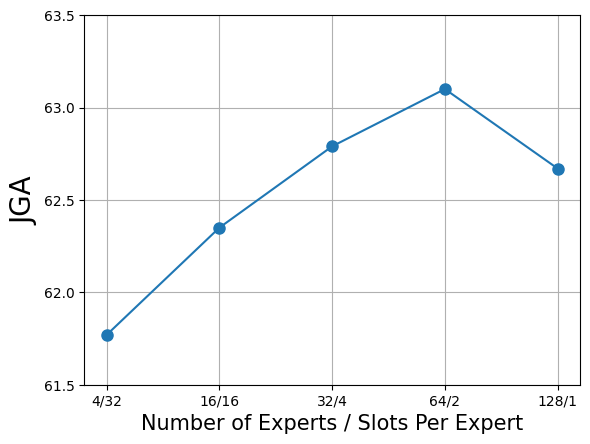}
  \label{fig:gnn}
  \end{subfigure}%
\begin{subfigure}{.25\linewidth}
  \centering
  \includegraphics[width=\linewidth]{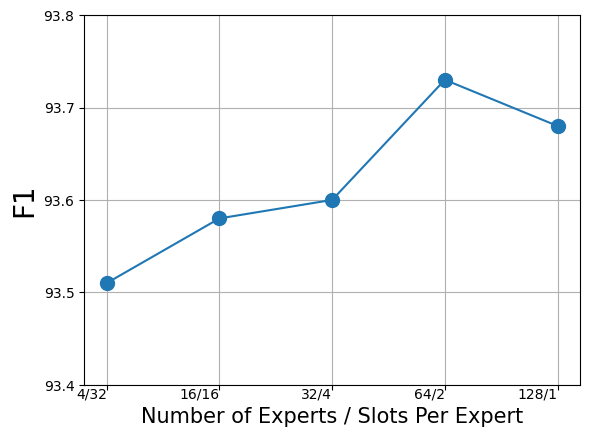}
  \label{fig:hidden}
  \end{subfigure}%
  \begin{subfigure}{.25\linewidth}
  \centering
  \includegraphics[width=\linewidth]{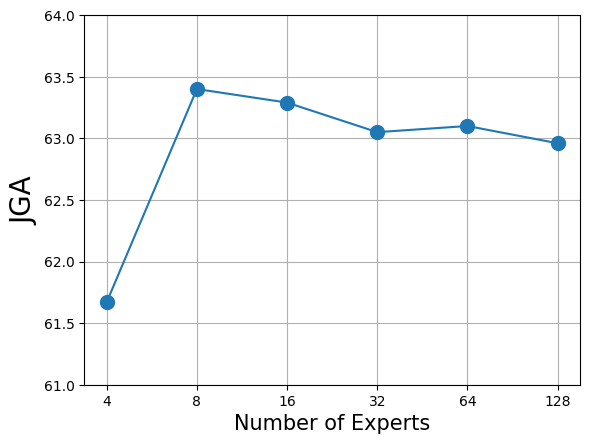}
  \label{fig:hidden}
  \end{subfigure}%
  \begin{subfigure}{.25\linewidth}
  \centering
  \includegraphics[width=\linewidth]{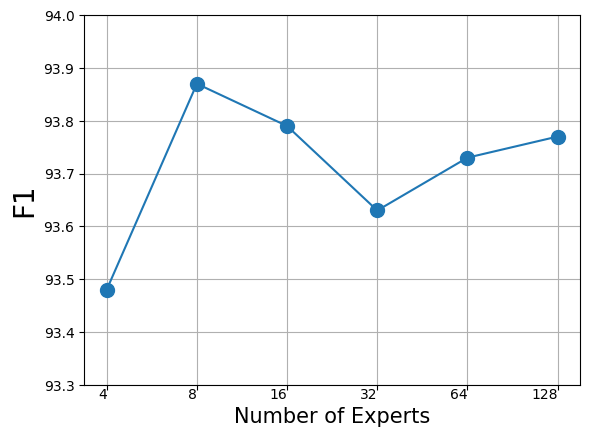}
  \label{fig:hidden}
  \end{subfigure}%
 \caption{(\textbf{Left}) Performance of SMETOD as a function of the number of experts, for models with a fixed number of experts $\times$ slots-per-expert. (\textbf{Right}) Performance of SMETOD trained with increased experts and 2 slots per expert. JGA and F1 scores are on MultiWOZ 2.1 dev set for DST.}
  \label{fig:architect1}
\end{figure*}

\subsubsection{Evaluation Results}
\footnotetext[6]{\url{https://github.com/budzianowski/multiwoz}}
SMETOD performs the best in terms of \textit{Inform} and \textit{Success} metrics on MultiWOZ 2.1, On MultiWOZ 2.2, it achieves new state of the art of \textit{Inform} and \textit{Combined} score, showing its overall strong performance.
The high \textit{Inform} rate indicates that entities and values generated by SMETOD are more factually correct.
Furthermore, we hypothesize that metrics hinder each other from being improved together and may require a mechanism to promote performance towards specific metrics, for example, REINFORCE~\cite{sutton1999policy}. 
Besides, we observe that only replacing the Feed-Forward layer in Transformer encoders as in~\cite{puigcerver2023sparse} without copying weights to experts doesn't generate the best results in our dialogue system.
It might be because their implementation requires a large amount of data to pre-train, which is inappropriate in the task-oriented scenario.
It demonstrates that by duplicating pre-trained weights and fine-tuning, SMETOD optimizes well for DST and NLG, respectively, maintaining the prior knowledge learned from the pre-trained model.

\begin{table}[h]
    \centering
    \begin{tabular}{l|c|c}
    \toprule
         \multirow{2}{*}{Model}& Small & Base\\
         & Time$\downarrow$ / Model Size&  Time$\downarrow$ / Model Size\\
         \midrule
         \midrule
         PPTOD& $1.000\times$ / $1.0\times$&$1.000\times$ / $1.0\times$\\
         TOATOD& $1.116\times$ / $1.1\times$&$1.113\times$ / $1.2\times$\\
         SMETOD& $1.005\times$ / $3.5\times$&$0.978\times$ / $4.0\times$\\
    \bottomrule
    \end{tabular}
    \caption{Comparison of the inference time with small and base-size models of PPTOD and TOATOD for NLG on MultiWOZ 2.1. All models are experimented with 5 same and randomly sampled batch sizes. The average time is reported. $\downarrow$: Smaller is better.}
    \label{tab:time}
\end{table}

\subsection{Time Complexity Analysis}
\label{sec:time}

We show in Table~\ref{tab:time} that SMETOD could make inferences without bringing about much extra time. 
SMETOD$_{small}$ is 3.5 times larger than PPTOD and TOATOD while achieving a similar inference speed as the former.
Our SMETOD$_{base}$ has even less inference time while its model size is 4 times of PPTOD$_{base}$.
It proves that we can achieve much better scaling while cost is roughly constant, with the benefit of improved performance.

\subsection{Impact of Expert Numbers}
\label{sec:architect}
We investigate the impact of expert and slot numbers in our models on the development set of MultiWOZ 2.1 for DST as illustrated in Figure~\ref{fig:architect1}.
First, we fix the total number of slots to 128 and vary expert numbers \{4, 16, 32, 64, 128\} by scaling slot numbers per expert.
Results suggest the best configuration is 64 experts and 2 slots per expert.
Then, we set the number of slots per expert to one and evaluate performance in terms of the number of experts. 
The number of experts 8 and 16 perform better than others.
It should be mentioned that the model size only scales with increasing expert numbers.
Meanwhile, we observe performance is not always increasing with the number of experts, indicating there is a trade-off between model size and the amount of training data.

\section{Conclusion}
 We propose an efficient fine-tuning approach based on Soft-MoE to satisfy requirements on diverse outputs in task-oriented dialogue systems.
 We demonstrate that incorporating Soft-MoE to our dialogue system achieves remarkable success on MultiWOZ baselines and optimizes outputs of each subproblem, showing it powerful technique for task-oriented dialogue systems with better scaling performance while maintaining time efficiency.

\appendix

\section*{Ethical Statement}

Using public dialogue benchmarks introduces the potential for biases stemming from the data collection method. Models trained on such datasets might encounter challenges when attempting to generalize to real-world scenarios or specific domains, as the data may not accurately represent these situations. Additionally, public dialogue datasets frequently lack essential context or metadata, rendering it difficult to comprehend the circumstances surrounding the conversations.

In our approach, we also rely on open-source code repositories. However, these repositories can present issues related to security vulnerabilities and compatibility. Furthermore, incomplete documentation can pose additional hurdles for further development. Given the absence of reliable support or comprehensive documentation, these factors can impede troubleshooting and hinder the overall development process.

\bibliographystyle{named}
\bibliography{ijcai24}

\end{document}